# Text and Audio Simplification: Human vs. ChatGPT

Gondy Leroy, Ph.D.[1], David Kauchak, Ph.D.[2], Philip Harber, M.D.[1], Ankit Pal[1], Akash Shukla[1],

[1]University of Arizona, Tucson, AZ; [2]Pomona College, Claremont, CA

**Abstract**

*Text and audio simplification to increase information comprehension are important in healthcare. With the introduction of ChatGPT, evaluation of its simplification performance is needed. We provide a systematic comparison of human and ChatGPT simplified texts using fourteen metrics indicative of text difficulty. We briefly introduce our online editor where these simplification tools, including ChatGPT, are available. We scored twelve corpora using our metrics: six text, one audio, and five ChatGPT simplified corpora (using five different prompts). We then compare these corpora with texts simplified and verified in a prior user study. Finally, a medical domain expert evaluated the user study texts and five, new ChatGPT simplified versions. We found that simple corpora show higher similarity with the human simplified texts. ChatGPT simplification moves metrics in the right direction. The medical domain expert's evaluation showed a preference for the ChatGPT style, but the text itself was rated lower for content retention.*

**Introduction**

Text readability has been recognized as an important topic in healthcare for decades. However, writing good and informative text is difficult. It becomes more difficult if the topic is complicated and its presentation must be simple enough to be accessible to readers with diverse educational backgrounds. There are two common approaches today: manual simplification, which mostly relies on readability formulas, and automated simplification, which is becoming available, but remains largely untested, with large language models (LLMs) such as those provided through ChatGPT.

Manual text simplification commonly relied on the use of a few readability formulas, e.g., the Flesch-Kincaid[1, 2] formula, and an expert rewriting a text. These usually provide a single number assigned to an entire text that is intended to represent its difficulty level. The formulas have been used for decades and continue to be utilized for evaluation today. For example, Hunt et al.[3] evaluated patient information leaflets with dermatology information; Meade and Dreyer evaluated consent forms[4]. Most of the formulas rely on intuition, for example, sentence and word length factor heavily into their outcomes. They have not been updated for the different types of media currently in use, e.g., email, or for other modes of information sharing, e.g., audio. They also do not provide concrete guidance on how to improve the text and have not been shown to predict or explain comprehension and retention of information.

Increasingly, different tools and metrics are becoming available for text simplification, e.g., Coh-Metrix[5] provides a wealth of metrics. Before ChatGPT became readily available, automated methods were being developed using machine learning algorithms to rewrite text. They often rely on metrics that can be efficiently applied to large corpora to evaluate outcomes but do not provide writer assistance at the document level. Evaluations with representative users are rare. Ondov et al.[6] reviewed 45 recent publications and found that 32 described current tools or methods, but only nine described their impact on human comprehension. This may be partially explained by the need to focus on the algorithms first. Only when algorithms have achieved high enough quality in conducting translations from difficult to simple will user studies be valuable. Many approaches still generate errors, which is unacceptable when distributing healthcare information. Devaraj et al.[7] reviewed factual errors in text simplification and found that most errors are due to the deletion of information. They also showed how many automated evaluation metrics are insensitive to measuring errors. This lack of good, automated metrics for evaluating simplification is problematic since many approaches rely on these metrics for the validation of the models. For example, Phatak et al.[8] use deep learning, readability formulas, and ROUGE[9] to validate their models. None of these metrics can replace user studies.

We focus on the intermediate approach and use automated suggestions for manual rewriting of the text. We have systematically evaluated different lexical, grammatical[10], and cohesion level features, discussed below, for their impact on text readability and understanding of the information contained in the text.  We use manual and machine learning approaches to discover these features that affect text and audio difficulty. Features that distinguish between simple and difficult text and that can also be used to automatically generate simplification suggestions are integrated into our online editor and visualization tool. We believe our approach is unique in three ways. First, our editor turns study findings into practical suggestions to simplify the text. Second, we visualize the text features and make it possible to visually compare different types of text. Third, we are including audio simplification in our studies and editor. We are now also adding ChatGPT simplification as a new component.

We present a comparison of different corpora at different difficulty levels and corpora simplified by ChatGPT. We use our experimentally verified metrics to compare the texts. We also compared the metrics of the corpora using cosine similarity with four texts simplified as part of an earlier user study with laypersons as readers [11]. We conclude with a medial domain expert evaluation of the content and style aspects of the texts simplified for the user study and now by ChatGPT using five different prompts.

**Online Editor**

Our prototype editor can be used online[1]. Our goal is to simplify and optimize content for reading and listening for a general audience. The current prototype does not use specific patient information to personalize information. Figure 1 shows the different components. There are five tabs that have the following functions:

- The "Simplification" tab allows a writer to upload or copy text, process the text, and review simplification suggestions by our editor. Each feature has been empirically validated. In addition, writers can choose to use ChatGPT to simplify, which is a new addition to our editor and is provided as-is.
- The "Lexical Chains" tab shows the spread of topics throughout a text. Each chain represents a topic and is assigned a color to show where it appears in the text. Three types of chains use increasingly more liberal synonym matching and so reflect the broader scope of terminology that is assigned to a topic. They rely on exact matches of terms, synonyms of terms, and semantically related terms to form chains.
- The "Statistics" tab shows the metrics related to text difficulty for the simplified text currently in the editor compared to the original, starting text.
- The "Speech" tab allows a user to create audio for the given text. We use Amazon Polly[2]. Users can choose the gender and voice accent for a text as well as add emphasis, and pauses, use a soft voice, and adjust volume, speech rate, and timbre for words and phrases in the text. Additional features include the ability to recognize dates and phone numbers in the text and pronounce them appropriately in the audio. Finally, specific pronunciations for words and adding full names for abbreviations can also be added. We are conducting studies to verify which text features affect audio comprehension. As more features to create audio become available, we add them.

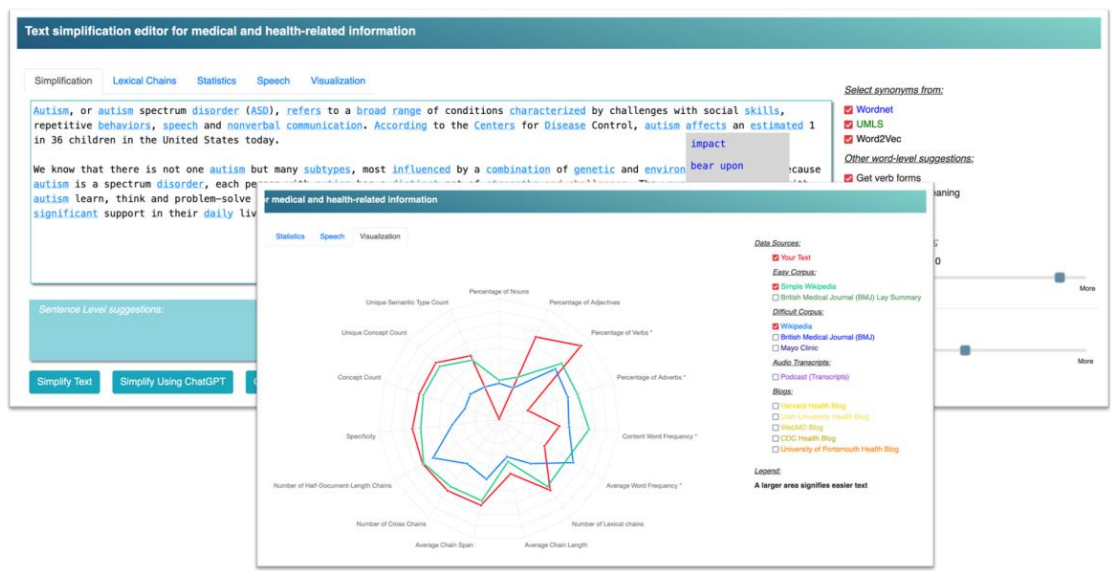

**Figure 1.** Overview of Text Editor Functionality

- The "Visualization" tab shows the text characteristics of the text and allows comparison with different existing corpora that show different difficulty levels. A writer can select or deselect any of these corpora. We use a radar graph with fifteen metrics where related metrics are shown in proximity to each other. The text a writer has been working on can be shown on the same radar graph. The scores on each metric are normalized (with the highest numbers of each corpus shown used to normalized) and axes are defined so that a higher number indicates higher readability. The numbers on the graph are dynamically normalized for the corpora shown. The axes of each feature are visualized such that a larger area on the graph indicates an easier text.

---
[1] http://simple.cs.pomona.edu:3000/
[2] https://aws.amazon.com/polly/

**Methods**

*Simplification Metrics*

In prior work, we conducted a series of user and machine learning studies leading to a variety of metrics that are associated with text difficulty and resulting comprehension. We use those metrics to compare different corpora in the study below. They can also be used online by users to evaluate their text and compare it with the corpora (Figure 1).

The first group of metrics provides measures of text difficulty by using statistics based on large corpora of "normal" text:

- Content word frequency[11, 12] shows how common words are based on Google Web Corpus[13] frequencies. The more common or frequent a word is, the easier it is to understand. This is in contrast with many readability formulas which assume that word length is the deciding factor. However, long words can be easier than short words when they are more frequently used, for example, 'diabetes' is more common and easier to understand than 'cyst'. We use the average frequencies in our analyses.
- Grammar frequency shows how common the underlying grammatical structures are in the sentences[14]. Using more common structures makes text easier to read.

We also show metrics reflecting specialized medical content[15]:

- Specificity shows how specific the terms are in the medical domain. We use Medical Subject Headings (MeSH[16]) with terms deeper in the hierarchy considered to be more specific and found to be more difficult.
- Ambiguity[17] reflects the vagueness of terms. Terms are looked up in the Unified Medical Language System (UMLS[18]) and their concepts retrieved. We then count how many different concepts are associated with that term. Terms that are associated with more different concepts reflect a higher ambiguity.
- Concept Density[17] reflects the number of unique concepts that are mentioned in a text. Having more different concepts reflects higher information density and so more difficult text. We use the UMLS to find the concepts associated with a term.
- Topic Density[17] is a broader metric where we use the number of unique semantic types associated with terms in the UMLS in a text. The same principle applies to concepts: more different semantic types in a text reflect higher information density and so more difficult text.

Surface text features related to parts-of-speech (POS) have also been associated with readability. We present nouns, verbs, adjectives, and adverbs which are content bearing and found in prior work to relate to difficulty[12]. We do not list other POS such as determiners, auxiliaries, and function words.

- Percentage of nouns: a higher percentage of nouns is associated with more difficult text.
- Percentage of adjectives: similar to nouns, a higher percentage is associated with more difficult text.
- Percentage of verbs: in contrast to nouns, a higher percentage of verbs is associated with easier text.
- Percentage of adverbs: similar to verbs, a higher percentage is associated with easier text.

We also calculate the number of topics and how they are spread throughout the text using lexical chains and crossing chains, which are associated with increased difficulty[19]:

- Average lexical chain count reflects how many different topics there are in a text, having more topics increases difficulty.
- Average lexical chain length reflects the number of repetitions of the content words. Easier texts had longer chain lengths (but not crossing), representing more repetition of content words.
- Average lexical chain span reflects across how much of the text the topic is spread. Again, easier texts have chains with longer spans (but not crossing).
- Average cross chains reflect how many topics overlap each other, more overlap results in more difficult text.

*Human and Artificial Corpora*

We use six corpora representing very different human-created text with different levels of difficulty. When automated collection was possible, larger corpora were created. However, we believe even with a smaller set, we capture the nature of the corpus sufficiently.

We collected webpages from the Mayo Clinic website (N=95) describing a variety of medical conditions. We also collected web pages from English Wikipedia (N=587) and Simple English Wikipedia (N=294) discussing medical conditions. To include more difficult text, we collected articles from the British Medical Journal (BMJ, https://ard.bmj.com) (N=668) as well as lay summaries (N=243) which can be considered the simpler versions. Finally, we also add transcripts of podcasts (N=539, https://healthliteracy.com, https://letstalkaboutmentalhealth.com.au, https://www.scientificamerican.com, http://www.cmhsrp.uic.edu, https://www.einstein.edu, https://tradeoffs.org). These are texts that represent audio because they were created as audio first.

We compare these corpora with five artificially generated corpora. To avoid creating text from scratch, since large language models (LLMs) have been shown to generate random content and errors, we used a set of 100 text snippets that patients might use when searching for health information. Specifically, we did a Google search on all concepts from the UMLS Metathesaurus that were tagged as either "disease" or "syndrome" (~10K terms). For many of these terms, the search results included a "People also ask" subset of results with a common question and an associated answer, along with a link. We randomly selected 100 snippets from these question/snippet responses. These snippets were chosen to represent what an average online health information consumer might encounter for such different conditions. We then simplified each of the snippets using ChatGPT 3.5 using the following prompts:

- Simplified: "Simplify the following text: …"
- Easier to understand: "Make the following text easier to understand: …."
- For ESL (English as a Second Language): "Change the following text to be easier to understand for someone who is a non-native English speaker: …."
- For older: "' Change the following text to be easier to understand for someone who is older: …"
- To read out loud: "Make the following text easier to understand when read out loud: …."

ChatGPT 3.5 was used since this was the newest version in production available when the study was conducted.

*Corpora Comparisons*

First, we calculate all metrics listed above for the twelve different corpora and present these as a descriptive overview. Second, we compare these corpora with text simplified from a previous study. In the previous study, a health educator working at a local community health center simplified four texts. The texts were part of a user study that helped verify our metrics and the usability of our online editor[20]. We refer to these four texts as the "user study text". These simplified text versions were better understood than the original versions by a lay audience. We use cosine similarity based on the normalized scores for the different metrics to show the similarity between the user study text and the corpora. Normalization of the score was done using the maximum value for each metric among all corpora.

*Domain Expert Evaluation*

A medical domain expert compared human and ChatGPT simplification. Our starting point is the four texts from the user study. We used ChatGPT to simplify the original texts using the different prompts. The simplified texts were then shown in random order per topic to the domain expert who was blinded to the conditions.

Each text was evaluated on four scales using a 1-5 rating with 1 being the worst and 5 being the best score. The scales used were: Content retention, Focus (capturing the main points), readability, and accuracy. A fifth scale focused on the target audience that would fit this simplification ranging from 1 to 5, with 1 representing technical expert (worst), and 5 a lay public (best).

**Results**

*Human and Artificial Corpora*

For each corpus, we have calculated the features associated with the texts (Table 1 and Table 2). Most results are as expected in comparison to our earlier work [12, 14, 17, 21]. Content word frequency is higher for lay BMJ text compared to the BMJ articles. Similarly, content word frequency is higher for simple Wikipedia compared to normal Wikipedia pages. Podcast content word frequencies are the second highest and only simple Wikipedia uses more common content words. The content word frequency of the Mayo Clinic text is lower, i.e., not as common, but this may be due to readability formulas being used as the main criteria and these emphasize short words and not necessarily common words.

Grammar frequency shows clearly how simple Wikipedia and lay summaries use the easiest grammar structure. Surprisingly, text from the Mayo Clinic website uses uncommon grammatical structures. This may again be due to the nature of the site. The text uses many bulleted lists which are good visual indicators but may distort the sentence grammar.

**Table 1.** Text Readability Features for our Corpora created by Human Writers

| | Human-Created Corpora | | | | | |
|---|---|---|---|---|---|---|
| Features | Mayo Clinic (N = 95) | Wikipedia Diseases (N = 294) | Simple Wikipedia Diseases (N =587) | BMJ (N = 668) | BMJ Lay Summaries (N = 243) | Podcasts (N = 539) |
| Word Count | 2048 | 318 | 315 | 394 | 211 | 2461 |
| Ordinariness | | | | | | |
| Content Word Frequency | 289,687,748 | 345,809,104 | 518,756,168 | 225,145,121 | 356,758,256 | 392,207,129 |
| Grammar Frequency | 3,117 | 8,020 | 7,019 | 5,229 | 8,735 | 4,059 |
| Healthcare Domain Specialty (Averages) | | | | | | |
| Specificity | 0.90 | 0.87 | 0.97 | 0.79 | 0.88 | 0.61 |
| Ambiguity | 0.14 | 0.17 | 0.14 | 0.24 | 0.20 | 0.07 |
| Concept Density | 0.27 | 0.31 | 0.26 | 0.48 | 0.38 | 0.12 |
| Topic Density | 0.04 | 0.04 | 0.08 | 0.12 | 0.14 | 0.02 |
| Parts-Of-Speech Features (%) | | | | | | |
| Nouns | 34 | 32 | 30 | 37 | 31 | 23 |
| Verbs | 16 | 14 | 17 | 13 | 17 | 19 |
| Adjectives | 9 | 12 | 9 | 13 | 11 | 9 |
| Adverbs | 3 | 4 | 5 | 3 | 4 | 6 |
| Topic Spread (Averages) | | | | | | |
| Lexical Chains | 0.060 | 0.055 | 0.056 | 0.059 | 0.052 | 0.033 |
| Lexical Chain Length | 0.002 | 0.010 | 0.011 | 0.009 | 0.015 | 0.002 |
| Lexical Chain Span | 0.452 | 0.460 | 0.464 | 0.392 | 0.534 | 0.490 |
| Lexical Cross Chains | 0.0600 | 0.0550 | 0.0562 | 0.0588 | 0.0518 | 0.0333 |

The domain specialty features as well as the topic spread features show mostly small differences. However, we have found earlier that they distinguish significantly between easy and difficult text when combined[15]. The podcast text is the most different from other sources. This is expected since this text is a transcript and was made for audio. The most notable differences are that more general terms are used (lower specificity), lower ambiguity of terms, and a lower number of topics are discussed per text.

The percentages of nouns and verbs follow the expected distribution, compared to our prior work, with the highest percentage of nouns being used in the most difficult text and the highest percentage of verbs in easier text. Again, podcast transcripts show the most different numbers. They use the lowest percentage of nouns (easiest) and the highest percentage of verbs (easiest).

The spread of topics in a text is very similar between similar corpora. Podcasts stand out again as having the least average number of chains, with chains that are the shortest, and the lowest number of crossing chains (i.e., overlapping topics).

The artificially generated corpora show similarities with the human corpora (Table 2). After simplification, the content word frequency increases for every simplification indicating that more common (i.e., easier) terms have replaced the more difficult terms. Grammar frequency increases when instructed explicitly to simplify or to create a version to read out loud. However, the grammar frequencies are lower (i.e., becoming more difficult) for simplifications targeted at ESL speakers and older people.

There is very little change in the domain specialty metrics. They slightly go down for all simplifications, i.e., somewhat less technical terms and lower concept density. However, in contrast to grammar frequency, the change is more pronounced for the simplifications targeted at ESL speakers and older people.

The parts-of-speech features show little variety, but the changes are in the direction of what we know to be easier. There is barely any change in the percentage of nouns and adjectives. There is slightly more change in the percentage of verbs. A somewhat larger change is seen with more specific prompts, i.e., for ESL, for older, and to read out loud.

The topic spread changes little, as can be expected with short text snippets.

Note that the corpora Simplified and Easier to Understand were generated with different prompts, but the outcomes were nearly identical.

**Table 2.** Text Readability Features for our Corpora created by ChatGPT.

|  | Artificially Created Corpora | | | | | |
|---|---|---|---|---|---|---|
| **Features** | Original snippet (N=100) | Simplified (N=100) | Easier to understand (N=100) | For ESL (N=100) | For older (N=100) | To read out loud (N=100) |
| Word Count | 156 | 181 | 174 | 174 | 176 | 156 |
| Ordinariness | | | | | | |
| Content Word Frequency | 330,704,302 | 372,494,762 | 372,494,762 | 370,435,797 | 376,743,175 | 379,503,773 |
| Grammar Frequency | 8,500 | 10,236 | 10,236 | 8,396 | 8,030 | 8,737 |
| Healthcare Domain Specialty (Averages) | | | | | | |
| Specificity | 1.04 | 1.01 | 1.01 | 1.03 | 1.03 | 1.02 |
| Ambiguity | 0.54 | 0.52 | 0.52 | 0.48 | 0.47 | 0.50 |
| Concept Density | 0.54 | 0.52 | 0.52 | 0.48 | 0.47 | 0.50 |
| Topic Density | 0.26 | 0.25 | 0.25 | 0.23 | 0.22 | 0.24 |
| Parts-Of-Speech Features (%) | | | | | | |
| Nouns | 33 | 32 | 32 | 32 | 31 | 31 |
| Verbs | 14 | 15 | 15 | 16 | 16 | 16 |
| Adjectives | 3 | 4 | 4 | 4 | 4 | 4 |
| Adverbs | 13 | 12 | 12 | 11 | 11 | 12 |
| Topic Spread (Averages) | | | | | | |
| Lexical Chains | 0.056 | 0.058 | 0.057 | 0.054 | 0.055 | 0.056 |
| Lexical Chain Length | 0.019 | 0.017 | 0.017 | 0.017 | 0.017 | 0.019 |
| Lexical Chain Span | 0.456 | 0.447 | 0.462 | 0.402 | 0.443 | 0.456 |
| Lexical Cross Chains | 0.056 | 0.057 | 0.056 | 0.053 | 0.054 | 0.056 |

*Corpora Comparisons*

Table 3 shows the simplification metrics for the text from the user study. In that study, we showed the simplified texts were largely better understood by a lay audience [11, 20]. The metrics show how simplified text used more common words and more common grammar. The domain difficulty levels changed only slightly because the human editor was instructed not to remove information or add additional information. However, the differences reflect how the editor's suggestions resulted in less dense medical content as well as fewer nouns and more verbs in the text.

**Table 3.** Text Readability Features for Four User Study Texts.

|  | Asthma | | Liver Cirrhosis | | Pemphigus | | Polycythemia Vera | |
|---|---|---|---|---|---|---|---|---|
| **Features** | Original | Simplified | Original | Simplified | Original | Simplified | Original | Simplified |
| Text Ordinariness | | | | | | | | |
| Content Word Frequency | 280,834,273 | 327,044,769 | 313,157,350 | 347,837,287 | 325,116,425 | 393,023,079 | 276,426,270 | 286,907,512 |
| Grammar Frequency | 4,019 | 11,773 | 2,499 | 2,778 | 6,431 | 5,707 | 4,049 | 18,916 |
| Healthcare Domain Specialty (Averages) | | | | | | | | |
| Specificity | 0.89 | 0.94 | 0.94 | 0.97 | 1.16 | 0.98 | 0.96 | 0.97 |
| Ambiguity | 0.22 | 0.17 | 0.28 | 0.22 | 0.26 | 0.21 | 0.30 | 0.27 |
| Concept Density | 0.42 | 0.33 | 0.65 | 0.50 | 0.50 | 0.40 | 0.56 | 0.50 |
| Topic Density | 0.09 | 0.08 | 0.13 | 0.12 | 0.17 | 0.18 | 0.18 | 0.16 |
| Parts-Of-Speech Features (%) | | | | | | | | |
| Nouns | 33 | 32 | 37 | 35 | 34 | 36 | 39 | 38 |
| Verbs | 13 | 9 | 14 | 11 | 18 | 17 | 12 | 14 |
| Adjectives | 15 | 16 | 13 | 15 | 12 | 12 | 11 | 13 |
| Adverbs | 4 | 3 | 4 | 4 | 2 | 3 | 5 | 5 |
| Topic Spread (Averages) | | | | | | | | |
| Lex. Chains | 0.052 | 0.059 | 0.053 | 0.049 | 0.060 | 0.073 | 0.060 | 0.055 |
| Lex. Chain Length | 0.006 | 0.006 | 0.007 | 0.008 | 0.019 | 0.020 | 0.018 | 0.019 |
| Lex. Chain Span | 0.265 | 0.263 | 0.391 | 0.422 | 0.590 | 0.497 | 0.405 | 0.426 |
| Cross Chains | 0.052 | 0.059 | 0.053 | 0.049 | 0.060 | 0.073 | 0.060 | 0.055 |

To reduce the number of comparisons and have the metrics represent multiple texts, we averaged the numbers for original and simplified texts (Table 4). These numbers are the basis for calculating cosine similarity with the corpora.

**Table 4.** Average Text Readability Features for Evaluated Texts.

| Features | User Study Original Texts | User Study Simplified Texts |
|---|---|---|
| Ordinariness | | |
| Content Word Frequency | 298,883,580 | 338,703,162 |
| Grammar Frequency | 4,250 | 9,794 |
| Healthcare Domain Specialty (Averages) | | |
| Specificity | 0.988 | 0.967 |
| Ambiguity | 0.264 | 0.219 |
| Concept Density | 0.532 | 0.435 |
| Topic Density | 0.140 | 0.136 |
| Parts-Of-Speech Features (%) | | |
| Nouns | 36 | 35 |
| Adjectives | 13 | 14 |
| Verbs | 14 | 13 |
| Adverbs | 4 | 4 |
| Topic Spread (Averages) | | |
| Lexical Chains | 0.056 | 0.059 |
| Lexical Chain Length | 0.013 | 0.013 |
| Lexical Chain Span | 0.413 | 0.402 |
| Lexical Cross Chains | 0.056 | 0.059 |

Table 5 shows the cosine similarity between the texts and the different corpora. We first compare the human-created corpora and how they relate to either the original or simplified user study texts (comparison within rows). We found that text from the Mayo Clinic is similar to both the original and the simplified text. For both versions of Wikipedia, we find they are more like the simplified text than the original. The scientific articles (BMJ) are more similar to the original text than the simplified one, while the reverse is true for the BMJ lay summaries. The podcasts show more similarities to the simplified text.

We are also interested in which corpus is most similar to either the original or simplified texts (comparison with columns). We find that the highest similarity to the original text was found for the BMJ texts and the lowest for the podcasts. When comparing with our simplified texts, we see the highest similarity with the BMJ lay summaries.

**Table 5.** Cosine Similarity between Corpora and the Original and Expert-Simplified Text.

| | User Study Original Texts | User Study Simplified Texts |
|---|---|---|
| **Human Created Corpora** | | |
| Mayo Clinic | 0.839 | 0.830 |
| Wikipedia | 0.866 | 0.899 |
| Simple Wikipedia | 0.846 | 0.878 |
| BMH | 0.900 | 0.885 |
| BMJ Lay Summaries | 0.891 | 0.923 |
| Podcasts | 0.816 | 0.842 |
| **Artificially Created (GPT) Corpora** | | |
| Original | 0.792 | 0.802 |
| Simplified | 0.795 | 0.819 |
| Easier to Understand | 0.797 | 0.821 |
| For ESL | 0.807 | 0.823 |
| For older | 0.808 | 0.822 |
| To read out loud | 0.799 | 0.816 |

For artificially simplified texts, we find that they all are more similar to the user study simplified texts than to the original texts (within row comparisons). Texts created for ESL speakers show the highest similarity to our study's simplified texts (within column comparisons).

Overall, the similarities with the original and simplified user study texts are lower for the artificial texts than the human-generated corpora.

*Domain Expert Evaluation*

The results of our medical domain expert evaluation are shown in Figure 2.

The user study's manually simplified text shows the best rating for the target audience, i.e., it is the most suitable for a lay reader. The artificially created simplifications show the best ratings in almost all categories except content retention. The prompt to make text easier to understand received, on average, has the best scores.

For content retention, the user study texts received one of the highest scores. This is not surprising since the health educator who simplified the texts for that study was instructed that no content could be deleted. The artificial texts could have content removed.

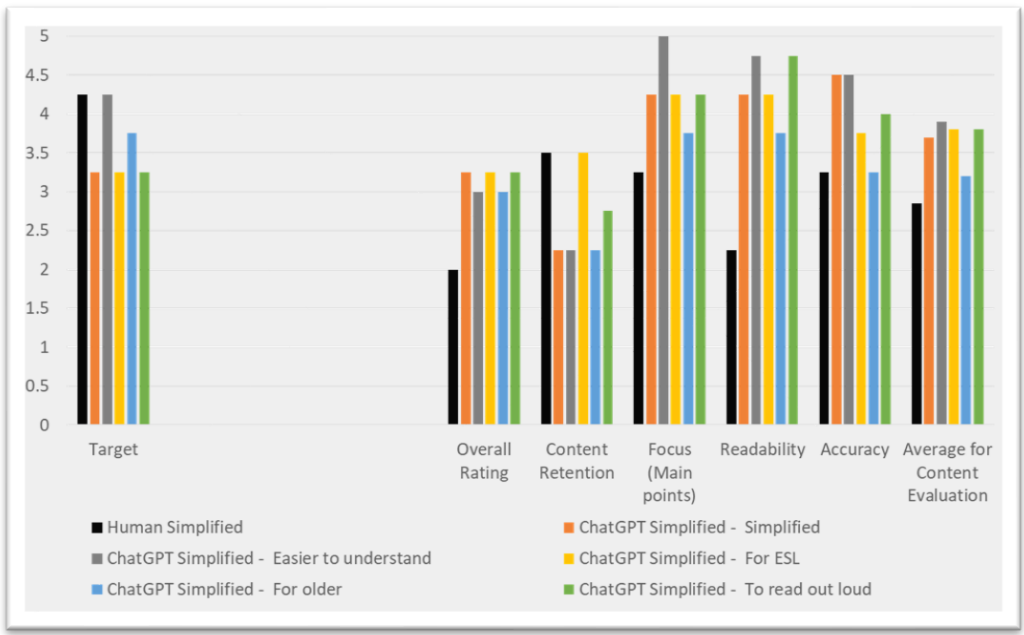

**Figure 2.** Domain Expert Evaluation

**Discussion and Conclusion**

To our knowledge, we are the first to compare and evaluate human and artificially simplified text using a wide variety of text features. In general, we find that for ChatGPT-simplified the text the simplified versions showed similar changes in the values of our metrics as human-simplified corpora. In addition to the descriptive analysis, we also compared artificial simplification with human simplification using cosine similarity and with a separate domain expert evaluation. Overall, the human corpora comparison results showed differences in metrics that were expected based on our prior work; easier information used more common words, more common grammar structures, fewer nouns but more verbs, and fewer crossing of topics. However, there were two exceptions. Text from the Mayo Clinic website showed unexpectedly different results for our features that are not in line with our earlier work. We believe this is due to the style of the pages and how they were optimized. They use spacing and bulleted lists to show chunks of information. They also were optimized using readability formulas. This naturally affects the writing style. Podcast transcripts are another exception. This corpus was included because these represent made-for-audio texts or conversations. As expected, they exhibit different features. In general, we find that ChatGPT simplified the text resulting in simplified versions that also reflected the same change in the values of our metrics as human simplified corpora.

Using texts from our prior user study as a benchmark we found a repetition of patterns, namely that known simpler corpora are more similar to our verified simple text. The same two corpora, Mayo Clinic text and podcasts, were exceptions. ChatGPT-generated texts showed lower overall similarities to our study texts.

Finally, a domain expert evaluation rated artificially generated simplification as preferable to manual. However, these versions reduced the total information content. As such, it is easier to present simpler text if difficult or detailed content can be removed. The implications for healthcare providers are significant: with the right checks and safeguards in place, automated simplification of text (for reading or audio creation) will be possible and efficient. More research is needed on how to ensure no errors are introduced (i.e., avoid hallucinations) and all necessary information remains present (i.e., avoid overgeneralization).

This study has several limitations. We worked with a single domain expert which allowed us to make comparisons between corpora. While the origin of the text was blinded to this expert, and the order randomized, we intend to engage with multiple experts and so improve the validity of our work. Furthermore, an extensive statistical comparison of all features between all corpora was beyond the scope of this work but will be included in future work where we will also have reader or listener comprehension scores. In addition, we limited our work to five prompts which we intended to reflect how healthcare providers, who are not prompt engineers, might use ChatGPT. Using more sophisticated prompts may bring different results. Finally, we focus on features we can automatically identify in text and for which we can automatically suggest improvements. We make no claims on any theory related to information comprehension.

In the future, we will continue adding features to simplify for a broad audience as we discover them through our studies and work on adding ChatGPT with content retention control. However, we increasingly focus on audio delivery of information. We will verify existing features that were important for text as well as focus on new audio-only features such as emphasis and pauses. We also intend to create an integrated metric that combines weighted features into one outcome for a text, similar to how cosine similarity was used in this paper. In addition, we envision a version of our editor that personalizes text and will be useful for provider-patient communication. For such a tool, a local LLM will be needed to avoid releasing protected and private information. Finally, we continue to develop our online interface to provide more user features, fix problems, and improve user-friendliness.

**Acknowledgments**

The research reported in this paper was supported by the U.S. National Library of Medicine of the National Institutes of Health under Award Number R01LM011975. The content is solely the responsibility of the authors and does not necessarily represent the official views of the National Institutes of Health.